\documentclass{article} 
\usepackage{iclr2027_conference,times}

\usepackage{amsmath,amsfonts,bm}

\def\eqref#1{equation~\ref{#1}}

\def\1{\bm{1}}

\DeclareMathAlphabet{\mathsfit}{\encodingdefault}{\sfdefault}{m}{sl}
\SetMathAlphabet{\mathsfit}{bold}{\encodingdefault}{\sfdefault}{bx}{n}

\usepackage{algorithm}
\usepackage{algpseudocode}
\usepackage{float}
\usepackage{wrapfig}
\usepackage{graphicx}
\usepackage{amsmath}
\usepackage{subcaption,adjustbox}
\usepackage{multirow}

\usepackage[utf8]{inputenc} 
\usepackage[T1]{fontenc}    
\usepackage{hyperref}       
\usepackage{url}            
\usepackage{booktabs}       
\usepackage{amsfonts}       
\usepackage{nicefrac}       
\usepackage{microtype}      
\usepackage{xcolor}         

\title{Tiny Recursive Policy: Recursive Latent \\ Refinement for Visuomotor Control}

\author{\\
\parbox{\textwidth}{
\centering
{\textbf{Hakan Aktas}$^{1}$}
\hspace{0.32em}
{\textbf{Cengiz Oztireli}$^{1}$} 
\hspace{0.32em} \\[0.6em]
\normalfont
$^{1}$ University of Cambridge, Cambridge, United Kingdom  
}}

\iclrfinalcopy

\begin{document}

\lhead{} 

\maketitle
\lhead{} 

\begin{abstract}
  Behaviour cloning has become a powerful approach for robotic manipulation, but many modern policies reconstruct their internal representation from a fixed observation window at each prediction and do not maintain persistent latent context throughout execution. We introduce \textbf{Tiny Recursive Policy (TRP)}, a parameter-efficient behaviour-cloning policy that combines persistent temporal state with recursive latent refinement. At each control step, TRP repeatedly updates high- and low-level latent states using a shared computation module before decoding an action sequence, and propagates the resulting state to subsequent predictions. We evaluate TRP across LIBERO-100, MetaWorld ML45, and the LIBERO-Mem benchmarks. TRP performs competitively across multi-task and few-shot settings and shows substantial gains on long-horizon and context-dependent tasks, while using significantly fewer parameters than large generative baselines. Execution-time perturbations show that reliance on the persistent carry increases with temporal-context demand, while training with carry perturbations substantially improves robustness to both carry and visual corruptions. Ablations reveal a strong interaction between persistent state and recursive refinement: the benefit of carry depends strongly on the amount of within-step computation, and the resulting performance cannot be explained by parameter count alone.
\end{abstract}

\section{Introduction}
\label{sec:intro}

Behaviour cloning (BC) has become a central paradigm for learning robotic manipulation policies from demonstrations. Recent advances have significantly improved its ability to model complex and long-horizon behaviours. Action-chunking methods predict temporally coherent sequences of future actions rather than individual control commands \citep{zhao2023learning}, while discrete latent-action approaches learn reusable representations of temporally extended behaviour \citep{lee2024behavior,mete2024quest}. Generative policies, particularly diffusion- and flow-based methods, have further improved continuous control through iterative action generation and expressive policy parameterisations \citep{chi2025diffusion,reuss2023goal,pan2026much}.

On the other hand, retaining temporal information is another important consideration for robotic policies. In long-horizon manipulation settings, the robot's actions may depend not only on the current observation, but also on the interaction history. Observation histories, action chunking, and temporal abstractions can capture this structure within a fixed temporal window, and many policies reconstruct their internal state from that window at each query. Recurrent models propagate latent context across action steps, allowing temporal interaction information to directly influence later actions and making the policy more robust than a fixed temporal window. This is especially relevant when visually similar observations occur at different stages of execution, where interaction history can provide additional context. This distinction becomes particularly important in partially observable settings, where identical or visually similar observations may require different actions depending on earlier interactions.

Furthermore, recurrence can provide more than temporal persistence. Rather than committing to an action sequence after a single forward pass, recursive refinement lets later computation revise and build on intermediate representations conditioned on the current observation and temporal model state. This provides a parameter-efficient way of increasing policy expressivity without increasing model size. Recent recursive reasoning models have demonstrated the effectiveness of this principle. Hierarchical Reasoning Model (HRM) \citep{wang2025hierarchical} performs repeated computation using interacting high- and low-level recurrent states operating at different timescales, while Tiny Recursive Model (TRM) \citep{jolicoeur2025less} shows that a small shared network can achieve strong reasoning performance through repeated latent refinement. These results motivate investigating whether recursive refinement, together with persistent temporal state, can provide an effective and parameter-efficient mechanism for robotic control.

In this work, we introduce \textbf{Tiny Recursive Policy (TRP)}, a behaviour-cloning policy that combines persistent temporal state with recursive latent refinement. At each control step, TRP repeatedly updates a pair of high- and low-level latent states using a small shared module before generating future actions. The resulting states are propagated to the next policy step, allowing subsequent predictions to incorporate temporal context. We additionally introduce a training-time carry perturbation objective that increases robustness to imperfect recurrent states. We evaluate TRP across multi-task, few-shot, long-horizon, and context-dependent manipulation settings on LIBERO, MetaWorld ML45, and LIBERO-Mem. TRP performs competitively across multi-task and few-shot benchmarks, while outperforming baselines on long-horizon and non-Markovian tasks. Through execution-time perturbations, we further show that reliance on the persistent carry increases with temporal-context demand: resetting the carry has almost no effect on MetaWorld but substantially degrades performance on LIBERO-LONG and LIBERO-Mem. Carry perturbation makes the resulting policy substantially more robust to both carry and observation corruptions. Finally, ablations over recursion depth, model size, and temporal gradient propagation show that TRP's gains arise from the interaction between persistent state and repeated computation, not parameter count alone.

\section{Related Work}
\label{sec:rel-work}
\paragraph{Behaviour Cloning}
Behaviour cloning (BC) learns policies from expert observation-action pairs and has long provided an effective framework for imitation learning \citep{pomerleau1988alvinn,bain1995framework,argall2009survey,osa2018algorithmic}. However, in robotic manipulation, demonstrations commonly exhibit multimodal action distributions and temporal associations, making direct single-step regression insufficient for many complex tasks. Early approaches addressed these issues using mixture-density policies \citep{mandlekar2021matters}, implicit energy-based policies \citep{florence2022implicit}, and discretised autoregressive action models such as Behaviour Transformer (BeT) \citep{shafiullah2022behavior} and VQ-BeT \citep{lee2024behavior}. Another line of work utilises temporal structure by generating extended action sequences or learning temporally abstracted latent skills. Action Chunking Transformer (ACT) predicts action chunks using a Conditional Variational Autoencoder (CVAE) \citep{zhao2023learning}, while latent-plan and skill-prior approaches compress temporally extended behaviour into reusable representations \citep{lynch2020learning,pertsch2021accelerating}. More recent approaches learn discrete latent action spaces, including PRISE \citep{zheng2024prise}, TAP \citep{jiang2022efficient}, H-GAP \citep{jiang2024h}, and QueST \citep{mete2024quest}, with QueST explicitly learning sequences of discrete skill tokens to improve multi-task and few-shot transfer.

Generative policies have become a dominant approach for continuous Behaviour Cloning. Diffusion Policy introduced conditional action diffusion for visuomotor control \citep{chi2025diffusion}, followed by goal-conditioned diffusion policies and extensions incorporating 3D observations, geometric equivariance, and larger policy architectures \citep{reuss2023goal,ze20243d,ke20243d,yang2024equibot,liu2025rdt}. Flow-based policies provide a related iterative formulation while enabling faster generation \citep{zhang2025flowpolicy,hu2024adaflow}. Although these approaches are commonly motivated by their ability to represent complex or multimodal action distributions, a recent work \citep{pan2026much} shows that much of their advantage can instead be attributed to stochasticity combined with supervised iterative computation; their Minimal Iterative Policy (MIP) matches flow-based policies using only two regression-based refinement steps.

\paragraph{Recursive Reasoning} Recent work on recursive reasoning models has explored iterative latent computation as an alternative to scaling model size or generating explicit chain-of-thought traces. Hierarchical Reasoning Model (HRM) \citep{wang2025hierarchical} introduced this direction using high- and low-level recurrent modules operating at different timescales, together with deep supervision and adaptive computation. Tiny Recursive Models (TRM) \citep{jolicoeur2025less} showed that much of HRM's architectural complexity was unnecessary, replacing the two-module hierarchy with a single small network that repeatedly refines a latent reasoning state and predicted solution, improving performance while substantially reducing the parameter count. This recursive-refinement principle has since been extended beyond discrete reasoning: Tiny Recursive Control (TRC) \citep{jain2026tiny} applies repeated prediction and trajectory simulation to optimal-control problems, while the Recursive Token Mapper (RTM) \citep{esmaeilzadeh2026one} replaces the latent mapper of style-based image generators with recursive refinement, improving image quality and distribution coverage. However, recursive refinement does not appear to transfer uniformly across settings: Autoregressive TRMs \citep{rauba2026tiny} show that TRM-style nested within-token refinement provides no advantage over standard recurrent Transformers, suggesting that its benefits may depend on the ability to repeatedly revise a candidate solution rather than committing autoregressively to individual outputs.

\section{Method}
\label{sec:method}

We now formalise TRP. The policy combines within-step recursive refinement with a latent carry propagated across control timesteps. Let the training dataset be composed of robot demonstration trajectories
\begin{equation}
\mathcal{D} = \{\tau^j\}_{j=1}^{N},
\qquad
\tau^j = \{(o_t^j, a_t^j)\}_{t=0}^{T_j-1},
\end{equation}
where $o_t^j$ denotes the robot observation at time $t$ in trajectory $j$, and $a_t^j$ denotes the corresponding action. The observation $o_t$ may contain visual observations, task descriptions, proprioceptive state or gripper state. At control time $t$, the model receives an observation window
\begin{equation}
O_t = \{o_{t-n_o+1}, \dots, o_t\},
\end{equation}
where $n_o$ is the observation horizon. The supervised prediction target is an action sequence
\begin{equation}
A_t = (a_t, a_{t+1}, \dots, a_{t+H_p-1}),
\end{equation}
where $H_p$ is the prediction horizon. The observation window is first encoded into a conditioning representation
\begin{equation}
x_t = f_{\mathrm{obs}}(O_t),
\end{equation}
where $f_{\mathrm{obs}}$ denotes the observation encoder. To predict a sequence of future actions, the conditioning representation is converted into a sequence of action-query embeddings,
\begin{equation}
X_t = \{x_t + p_h\}_{h=0}^{H_p-1},
\end{equation}
where $p_h$ is a learnable embedding indicating the future action index. Thus, the recurrent computation core operates over the future action slots rather than directly over the raw observation sequence. The core maintains two latent states,
\begin{equation}
c_t = (z_{H,t}, z_{L,t}),
\qquad
z_{H,t}, z_{L,t} \in \mathbb{R}^{H_p \times d},
\end{equation}
where $z_{H,t}$ is a high-level latent state and $z_{L,t}$ is a low-level latent state. The pair $c_t$ is the persistent carry. At the beginning of an episode, it is initialised as
\begin{equation}
c_0 = (z_H^{\mathrm{init}}, z_L^{\mathrm{init}}),
\end{equation}
and after each policy evaluation, it is updated and reused at the next timestep. Let $\mathcal{T}_\theta$ be a shared compute module. In practice, this can be implemented as a small transformer block, but the same block is reused for both low-level refinement and high-level update. We define an update operator
\begin{equation}
\mathcal{U}_\theta(z, u) = \mathcal{T}_\theta(z + u),
\end{equation}
where $z$ is the latent state being updated and $u$ is the conditioning input injected into the update. Given the incoming carry $c_t^{\mathrm{in}} = (z_{H,t}^{0}, z_{L,t}^{0,0})$, the recursive computation block performs $R$ high-level cycles. Each high-level cycle contains $K$ low-level refinement steps. The low-level state is carried between high-level cycles, such that $z_{L,t}^{1,0} = z_{L,t}^{0,0}$ and $z_{L,t}^{r,0} = z_{L,t}^{r-1,K}$ for $r>1$:
\begin{equation}
z_{L,t}^{r,\ell}
=
\mathcal{U}_\theta
\left(
z_{L,t}^{r,\ell-1},
z_{H,t}^{r-1} + X_t
\right),
\quad
z_{H,t}^{r}
=
\mathcal{U}_\theta
\left(
z_{H,t}^{r-1},
z_{L,t}^{r,K}
\right),
\end{equation}
where $\ell = 1,\dots,K$ and $r = 1,...,R$. The low-level state $z_L$ performs fast repeated refinement conditioned on the current high-level state and the observation. These equations describe the forward recursive computation. However, during training, we follow the truncated gradient scheme of \citet{jolicoeur2025less}: the first $R-1$ high-level cycles are evaluated without gradient tracking, and their resulting latent states are detached before the final cycle. Therefore, gradients propagate only through the $K$ low-level refinement steps and the high-level update of the final cycle. At inference time, all $R$ cycles are evaluated identically, as gradient tracking is not required.

\begin{algorithm}[t]
\caption{Stateful recursive policy step}
\label{alg-1}
\begin{algorithmic}[1]
\Require Observation window $O_t$, incoming carry $c_t^{\mathrm{in}}=(z_H,z_L)$
\Require Shared computation module $\mathcal{U}_\theta$, action head $h_\phi$
\Require Number of high-level cycles $R$, number of low-level refinement steps $K$
\State $x_t \gets f_{\mathrm{obs}}(O_t)$ \Comment{Encode current observation window}
\State $X_t \gets (x_t+p_0,\dots,x_t+p_{H_p-1})$ \Comment{Add future action index}
\For{$r = 1,\dots,R$}
\For{$k = 1,\dots,K$}
\State $z_L \gets \mathcal{U}_\theta(z_L, z_H + X_t)$
\Comment{Fast low-level refinement}
\EndFor
\State $z_H \gets \mathcal{U}_\theta(z_H, z_L)$
\Comment{Slow high-level update}
\EndFor
\State $\hat{A}_t \gets h_\phi(z_H)$
\Comment{Predict future action sequence}
\State $c_t^{\mathrm{out}} \gets (z_H,z_L)$
\Comment{Updated persistent latent carry}
\State \Return $\hat{A}_t, c_t^{\mathrm{out}}$
\end{algorithmic}
\end{algorithm}
After the recursive updates, the final states are decoded into an action sequence:
\begin{equation}
\hat{A}_t
=
h_\phi(z_{H,t}^{R}),
\end{equation}
where $h_\phi$ is the action head and
\begin{equation}
\hat{A}_t =
(\hat{a}_t, \hat{a}_{t+1}, \dots, \hat{a}_{t+H_p-1}).
\end{equation}

We experimented with different carry latent combinations to feed into the action head and observed that, while there isn't a large difference, using $z_H$ only generally yielded better performance than other options (concatenation, summation, $z_L$ only), so we use it as the default for all evaluations.

The output carry is defined as
\begin{equation}
c_t^{\mathrm{out}}
=
(z_{H,t}^{R}, z_{L,t}^{R,K}).
\end{equation}
This carry is passed to the next policy iteration:
\begin{equation}
c_{t+1}^{\mathrm{in}}
=
\operatorname{sg}
\left(
c_t^{\mathrm{out}}
\right),
\end{equation}
where $\operatorname{sg}(\cdot)$ denotes stop-gradient. Thus, the policy is stateful at inference time, since the latent states persist across timesteps, but during training, the carry is detached between consecutive timesteps, preventing gradients from propagating across policy evaluations while retaining recurrent state information in the forward pass. The action prediction objective is standard supervised behaviour cloning over action sequences:
\begin{equation}
\mathcal{L}
=
\mathbb{E}_{\tau \sim \mathcal{D},\, t}
\left[
\sum_{h=0}^{H_p-1}
\left||
\hat{a}_{t+h} - a_{t+h}
\right||_2^2
\right].
\end{equation}
Equivalently, the model learns a deterministic sequence policy
\begin{equation}
\hat{A}_t
=
\pi_\Theta(O_t, c_t^{\mathrm{in}}),
\end{equation}
where $\Theta = \{\theta,\phi\}$ together with the observation encoder parameters.

Optionally, the model can be trained with carry perturbation to improve robustness to imperfect latent states. Let $c_t^{\mathrm{out}}$ be the clean carry and let $\epsilon_c \sim \mathcal{N}(0,I)$. A perturbed carry can be formed as
\begin{equation}
\tilde{c}_t^{\mathrm{out}}
=
(1-\lambda)c_t^{\mathrm{out}}
+
\lambda \epsilon_c,
\end{equation}
where $\lambda$ controls the perturbation strength. A second supervised prediction from the perturbed carry gives an auxiliary loss,
\begin{equation}
\mathcal{L}
=
\mathcal{L}_{\mathrm{clean}}
+
\beta
\mathcal{L}_{\mathrm{perturbed}},
\end{equation}
where $\beta$ weights the auxiliary objective. This encourages the recurrent policy to remain stable when the persistent latent carry is noisy or imperfect, without changing the inference-time policy. The proposed policy is also summarised in Algorithm \ref{alg-1} which describes one stateful policy evaluation.

\begin{table}[!th]
    \centering
    \caption{\textbf{Multi-task and few-shot performance on LIBERO-100.}
Success rates are reported for LIBERO-90 and LIBERO-LONG. Few-shot models are
pretrained on LIBERO-90 and fine-tuned using five demonstrations per
LIBERO-LONG task. Parameters represent the backbone parameter count.}
    \label{tab:results}
    \begin{tabular*}{\textwidth}{@{\extracolsep{\fill}}lcccc}
        \toprule
        & \multicolumn{2}{c}{\textbf{Multi-task}}
        & \multicolumn{1}{c}{\textbf{Few-shot}}
        & \\
        \cmidrule(lr){2-3}
        \cmidrule(lr){4-4}
        \textbf{Model}
        & \textbf{LIBERO-90}
        & \textbf{LIBERO-LONG}
        & \textbf{LIBERO-LONG}
        & \textbf{Params (M)} \\
        \midrule
        ResNet-T \citep{liu2023libero}                         & 0.844          & 0.441          & 0.519          & 4.8   \\ 
        DP  \citep{chi2025diffusion}        & 0.916          & 0.849          & 0.688 & 336.3 \\ 
        ACT \citep{zhao2023learning}       & 0.704          & 0.500          & 0.240          & 13.87 \\
        VQ-BET  \citep{lee2024behavior}     & 0.724         & 0.538          & 0.370          & 6.43  \\ 
        Quest \citep{mete2024quest}        & 0.861          & 0.704          & 0.427          & 18.48 \\ 
        MIP \citep{pan2026much}            & 0.918          & 0.846          & 0.641          & 336.3 \\ 
        FLOW \citep{pan2026much}           & 0.899          & 0.897          & 0.558          & 336.3 \\
        \midrule
        TRP                               & 0.929 & \textbf{0.919} & 	\textbf{0.704}          & \textbf{4.2} \\ 
        TRP-perturbed                     & \textbf{0.935}          & 0.913          & 0.696        & \textbf{4.2} \\ 
        TRP-no-carry                      & 0.733          & 0.783          & 0.374          & \textbf{4.2} \\ 
        \bottomrule
    \end{tabular*}
\end{table}

\section{Experimental Results}
\label{sec:results}
We evaluate TRP across LIBERO-100 \citep{liu2023libero}, MetaWorld ML45 \citep{yu2020meta}, and LIBERO-Mem \citep{chung2026rethinking}. On LIBERO-100 and MetaWorld, we consider both multi-task learning and few-shot generalisation. For MetaWorld, we follow the ML45 evaluation protocol of \citet{mete2024quest}. The benchmark contains 45 training tasks and five held-out tasks. We collect 100 demonstrations per training task using the scripted policies provided by MetaWorld and evaluate each task over 50 rollouts. For few-shot evaluation, the multi-task model is fine-tuned using five demonstrations from each of the five held-out tasks. The reported results are over 3 seeds.

We evaluate three variants of our method: \textbf{TRP}, our full model with persistent latent carry across timesteps; \textbf{TRP-no-carry}, which discards the carry after each policy evaluation; and \textbf{TRP-perturbed}, which applies the carry perturbation objective during training. We compare against behaviour-cloning baselines from three families. Among policies that predict actions directly, \textbf{ResNet-T} \citep{liu2023libero} is the LIBERO reference policy, a ResNet encoder followed by a temporal transformer over recent observations; \textbf{LSTM-GMM} \citep{mandlekar2021matters} is a recurrent policy with a Gaussian-mixture action head and, like TRP, carries state across timesteps; and \textbf{ACT} \citep{zhao2023learning} predicts chunks of actions with a transformer trained as a conditional VAE. Two baselines act through discrete codes: \textbf{VQ-BeT} \citep{lee2024behavior} tokenises action chunks with residual vector quantisation and predicts the tokens with a GPT-style transformer, and \textbf{QueST} \citep{mete2024quest} learns a vocabulary of skill tokens with finite scalar quantisation and predicts them autoregressively. The remaining baselines refine actions iteratively: \textbf{Diffusion Policy} (DP) \citep{chi2025diffusion} denoises an action sequence over 100 DDPM steps with a 1-D convolutional U-Net, and \textbf{MIP} and \textbf{Flow} \citep{pan2026much} use the same U-Net with a two-step iterative update and rectified flow, respectively; DP's transformer variant (\textbf{DiT}) is also included in our model-size study. All models use the same vision encoder where possible for fairness.

\begin{figure}
    \centering
    \includegraphics[width=1\linewidth]{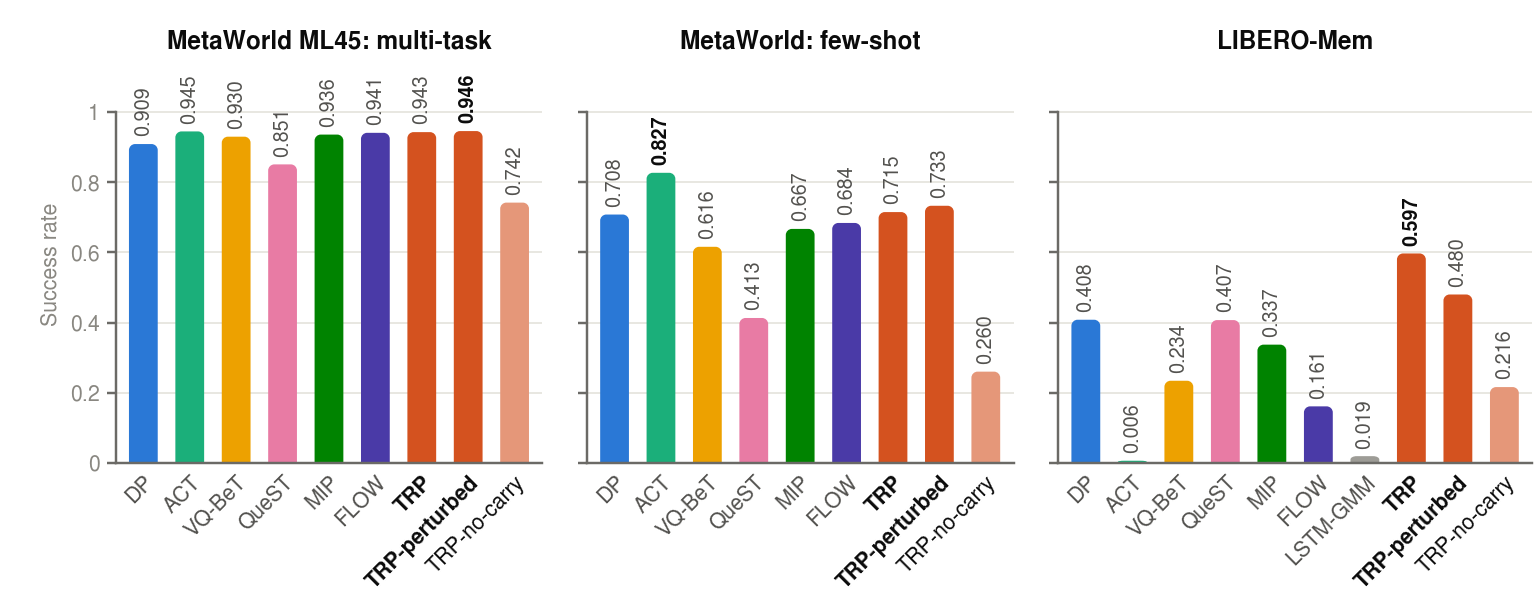}
    \caption{\textbf{Multi-task, few-shot, and context-dependent evaluation.}
Success rates on MetaWorld ML45 multi-task learning and five-shot transfer to
the five held-out ML45 tasks, together with subgoal completion on LIBERO-Mem.
MetaWorld follows the evaluation protocol of \citet{mete2024quest}. The
comparison highlights the effect of removing or perturbing the persistent
latent carry across increasingly transfer- and context-dependent settings.}
    \label{fig:mw-mem-results}
\end{figure}

\subsection{Multi-Task Learning}

We first evaluate whether TRP can learn a single policy across a diverse set of manipulation tasks. On LIBERO, models are trained using all 50 demonstrations for each task and evaluated over 50 rollouts. We report results on both LIBERO-90, which primarily contains shorter-horizon tasks, and LIBERO-LONG, which contains longer manipulation behaviours. We additionally evaluate on MetaWorld ML45 following the protocol of \citet{mete2024quest}.

As shown in Table~\ref{tab:results}, TRP performs strongly across both LIBERO multi-task settings. On LIBERO-90, TRP and TRP-perturbed outperform all baselines, with TRP-perturbed obtaining the highest success rate. The difference becomes clearer on LIBERO-LONG, where the margin over the strongest non-TRP baseline widens. The MetaWorld results in Figure~\ref{fig:mw-mem-results} show a similar regime. Multi-task ML45 performance is high for several methods, with ACT, MIP, Flow, and the two full TRP variants all concentrated near the top of the range. TRP-perturbed achieves the highest mean, but only by a small margin. In contrast, TRP-no-carry performs substantially worse despite using the same backbone and recursive computation. As examined further in Section \ref{subsec:ablations}, this gap arises primarily from training with persistent carry: wiping the carry from an already trained TRP has almost no effect on MetaWorld performance. Thus, on these tasks the carry shapes the learned policy during training without being required as online temporal context.

\subsection{Few-Shot Generalisation}

Next, we evaluate transfer to tasks with very limited task demonstration data. On LIBERO, models are first trained on the complete LIBERO-90 dataset and then fine-tuned using only five demonstrations from each LIBERO-LONG task. On MetaWorld, we again follow \citet{mete2024quest}: the model trained jointly on the 45 ML45 training tasks is fine-tuned using five demonstrations from each of the five held-out tasks. This setting tests whether representations learned from many source tasks can be adapted to new behaviours using limited data.

On LIBERO-LONG, TRP remains the strongest model in the five-shot setting, reaching 0.704 success, while TRP-perturbed performs similarly. The margins between the full TRP variants and the strongest generative baseline are relatively small, but the comparison with TRP-no-carry is much larger. Removing the persistent state reduces performance to roughly half that of the full model. Interestingly, this degradation is much stronger than in standard multi-task training, suggesting that the carried latent context is useful when the target behaviour is learned from very few examples. On the other hand, while TRP outperforms most of the strong baselines in the MetaWorld few-shot setting, ACT performs best in this setting. However, it should be noted that in the Libero few-shot setting, ACT is the worst performer among the baselines. The no-carry variant also falls short of every policy in this setting, supporting the importance of persistent model state.

\begin{table}[t]
\centering
\small
\setlength{\tabcolsep}{6pt}
\begin{tabular}{llcccc}
\toprule
 & & & \multicolumn{3}{c}{Execution-time perturbation} \\
\cmidrule(lr){4-6}
Benchmark & Model & Clean & Carry wiped & Blackout (short) & Blackout (long) \\
\midrule
\multirow{3}{*}{\shortstack[l]{MetaWorld ML45}}
 & TRP            & 0.943 & 0.941 & 0.659 & 0.515 \\
 & TRP-perturbed  & \textbf{0.946} & 0.942 & \textbf{0.672} & \textbf{0.531} \\
 & \textcolor{gray}{TRP-no-carry} & \textcolor{gray}{0.742} & \textcolor{gray}{--} & \textcolor{gray}{0.492} & \textcolor{gray}{0.347} \\
\midrule
\multirow{3}{*}{\shortstack[l]{LIBERO-Long}}
 & TRP            & \textbf{0.919} & 0.578 & 0.633 & 0.438 \\
 & TRP-perturbed  & 0.913 & \textbf{0.863} & \textbf{0.769} & \textbf{0.674} \\
 & \textcolor{gray}{TRP-no-carry} & \textcolor{gray}{0.783} & \textcolor{gray}{--} & \textcolor{gray}{0.524} & \textcolor{gray}{0.412} \\
\midrule
\multirow{3}{*}{\shortstack[l]{LIBERO-Mem}}
 & TRP            & \textbf{0.597} & 0.347 & 0.366 & 0.304 \\
 & TRP-perturbed  & 0.480 & \textbf{0.414} & \textbf{0.390} & \textbf{0.403} \\
 & \textcolor{gray}{TRP-no-carry} & \textcolor{gray}{0.216} & \textcolor{gray}{--} & \textcolor{gray}{0.136} & \textcolor{gray}{0.117} \\
\bottomrule
\end{tabular}
\caption{\textbf{Closed-loop performance under execution-time perturbations}. \emph{Clean} reports the corresponding headline result. \emph{Carry wiped} resets the recurrent carry before every prediction. \emph{Blackout} blanks all camera inputs for a short or long window mid-episode before restoring them (8/20 environment steps on MetaWorld and 16/32 on LIBERO).}
\label{tab:carry_perturbation}
\end{table}

\begin{table}[b]
    \centering
    \begin{subtable}[t]{0.56\textwidth}
        \centering
        \caption{Interaction between persistent carry and recursive refinement.}
        \label{tab:carry_recursion}
        \begin{adjustbox}{max width=\linewidth}
        \begin{tabular}{llccc}
            \toprule
            \textbf{Carry} & \textbf{Cycles} & \textbf{MetaWorld} & \textbf{LIBERO-LONG} & \textbf{LIBERO-Mem} \\
            \midrule
            on  & H3L6 & \textbf{0.943} & \textbf{0.919} & \textbf{0.597} \\
            on  & H1L1 & 0.630          & 0.206          & 0.362          \\
            off & H3L6 & 0.742          & 0.783          & 0.216          \\
            off & H1L1 & 0.794          & 0.642          & 0.305          \\
            \bottomrule
        \end{tabular}
        \end{adjustbox}
    \end{subtable}\hfill
    \begin{subtable}[t]{0.41\textwidth}
        \centering
        \caption{Temporal gradient propagation through the carry.}
        \label{tab:tbptt}
        \begin{adjustbox}{max width=\linewidth}
        \begin{tabular}{lcc}
            \toprule
            \textbf{Carry gradient} & \textbf{LIBERO-LONG} & \textbf{LIBERO-Mem} \\
            \midrule
            detached (TRP)  & \textbf{0.919} & \textbf{0.597} \\
            TBPTT, $k=4$    & 0.882          & 0.441          \\
            TBPTT, $k=8$    & 0.816          & 0.370          \\
            TBPTT, $k=16$   & 0.836             & 0.326          \\
            \bottomrule
        \end{tabular}
        \end{adjustbox}
    \end{subtable}
    \caption{\textbf{Ablations of the persistent latent carry.}
(a) compares the default recursive configuration with a single-cycle model,
with and without carry propagation. (b) compares the detached carry used by
TRP against truncated backpropagation through time (TBPTT).}
    \label{tab:carry_ablations}
\end{table}

\subsection{Evaluation on Context-Dependent Tasks}

We further evaluate on LIBERO-Mem \citep{chung2026rethinking}, a benchmark designed specifically for non-Markovian manipulation under partial observability. LIBERO-Mem contains 10 tasks spanning object motion, object sequence, object relations, and object occlusion. Unlike standard manipulation tasks, the current observation is intentionally insufficient to uniquely determine the correct action: visually similar states can require different behaviours depending on the preceding interaction history. Tasks include repeated manipulation of the same object, ordered interactions with multiple visually similar objects, and reasoning about objects after occlusion. LIBERO-Mem uses a subgoal-aware evaluation rather than relying only on binary task success, which we follow.

As shown in Figure~\ref{fig:mw-mem-results}, this setting produces the largest separation between TRP and the comparison methods. Standard TRP achieves a subgoal-completion score of 0.597, while the strongest non-TRP methods remain around 0.41. This contrasts with MetaWorld multi-task learning, where several methods achieve similar performance, and with the standard LIBERO benchmarks, where the margins are smaller. Consequently, TRP is relatively the most advantageous when successful execution depends on information not recoverable from the current observation alone.

The TRP variants provide further evidence for this interpretation. Removing the carry substantially reduces performance, while carry perturbation produces an intermediate result. This ordering is consistent with the persistent latent state carrying useful task context across policy evaluations. 

Strong action-generation methods also do not necessarily translate into strong performance in this setting. Diffusion Policy and QueST, for example, are competitive on several of the multi-task and few-shot evaluations but remain substantially below TRP on LIBERO-Mem. This suggests that expressive action distributions or temporal action abstractions alone do not fully substitute for state propagated across successive policy evaluations. Consequently, we view the LIBERO-Mem result as complementary to the long-horizon experiments: LIBERO-LONG tests whether a policy can handle extended execution, whereas LIBERO-Mem more directly tests whether information from earlier interaction can influence later interactions.

\subsection{Ablations} \label{subsec:ablations}

\paragraph{Execution-time role of the carry.} We first probe how trained policies use the persistent carry during closed-loop execution. Table~\ref{tab:carry_perturbation} compares each policy against two execution-time perturbations. In \emph{carry wiped}, we reset the carry to its initial value before each prediction, preventing information from propagating across policy evaluations while leaving the current observation unchanged. In \emph{blackout}, all camera inputs are blanked for a short or long interval mid-episode and subsequently restored. This removes visual information while also corrupting carry with uninformative observations. TRP-no-carry provides a stateless reference.

\paragraph{Carry reliance across benchmarks.} The effect of the carry differs sharply between training and execution. On MetaWorld ML45, wiping the carry from a trained TRP has almost no effect: TRP falls only from 0.943 to 0.941. In contrast, TRP-no-carry, which uses the same reset-before-prediction inference procedure but was trained without inter-step carry, reaches only 0.742. The resulting gap is therefore entirely a training-time effect rather than evidence that MetaWorld requires persistent temporal context at execution. One possible explanation is that propagating state during training provides a useful warm start for the recursive computation or improves the conditioning of the optimisation problem, even though the final policy does not depend on that state. This changes on the more temporally demanding benchmarks. Wiping the carry reduces standard TRP by approximately 37\% on LIBERO-LONG and 42\% on LIBERO-Mem, showing that the trained policy increasingly relies on propagated state during execution. Thus, carry appears to play two distinct roles: it can shape the learned policy during training even when dispensable at test time, and on temporally dependent tasks it also provides useful online context.

\paragraph{Robustness from carry perturbation.} Training with carry perturbations substantially reduces sensitivity to both carry removal and visual blackout on LIBERO-LONG and LIBERO-Mem. Wiping the carry causes only modest degradation for TRP-perturbed compared with standard TRP, and the same pattern holds under short and long blackouts. Under long blackout, for example, TRP-perturbed retains 0.674 success on LIBERO-LONG, compared with 0.438 for TRP, and also remains stronger on LIBERO-Mem. TRP-perturbed additionally degrades less than TRP-no-carry in these settings, suggesting that perturbation training produces a more robust recurrent state when visual evidence is temporarily unavailable. MetaWorld behaves differently: since wiping the carry has negligible effect, blackout primarily removes useful visual input and TRP and TRP-perturbed degrade similarly. These robustness gains come with little cost to clean performance, although standard TRP remains stronger on LIBERO-Mem.

\begin{figure}[t]
    \centering
    \begin{subfigure}[t]{0.46\textwidth}
        \centering
        \includegraphics[width=\linewidth]{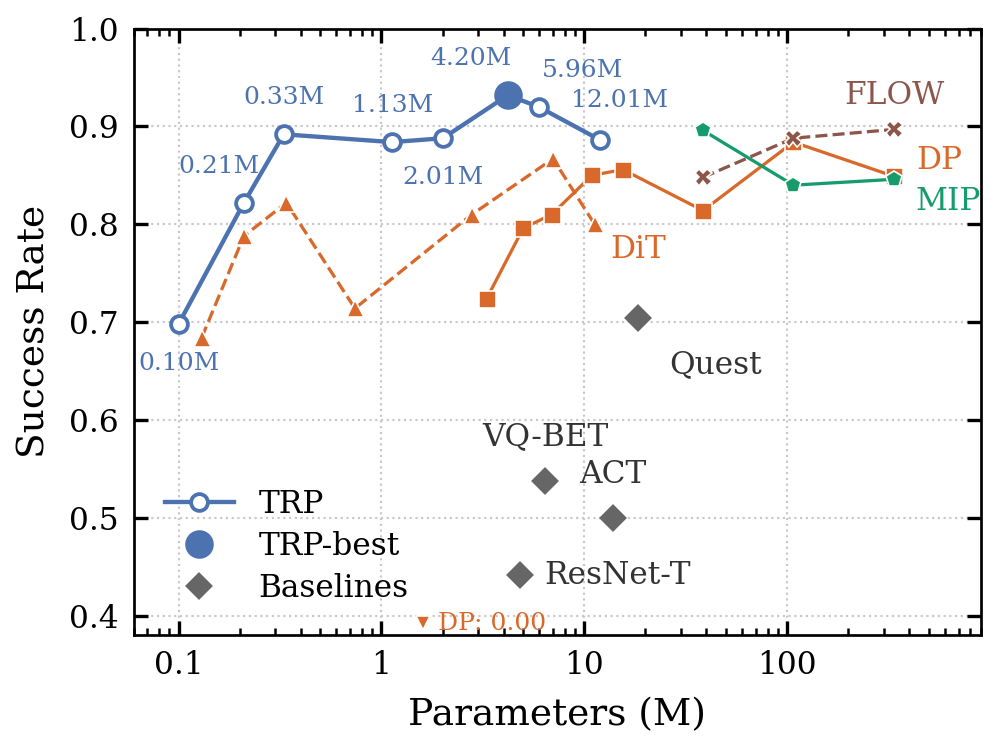}
        \caption{Model size.}
        \label{fig:size}
    \end{subfigure}%
    \hspace{0.05\textwidth}%
    \begin{subfigure}[t]{0.46\textwidth}
        \centering
        \includegraphics[width=\linewidth]{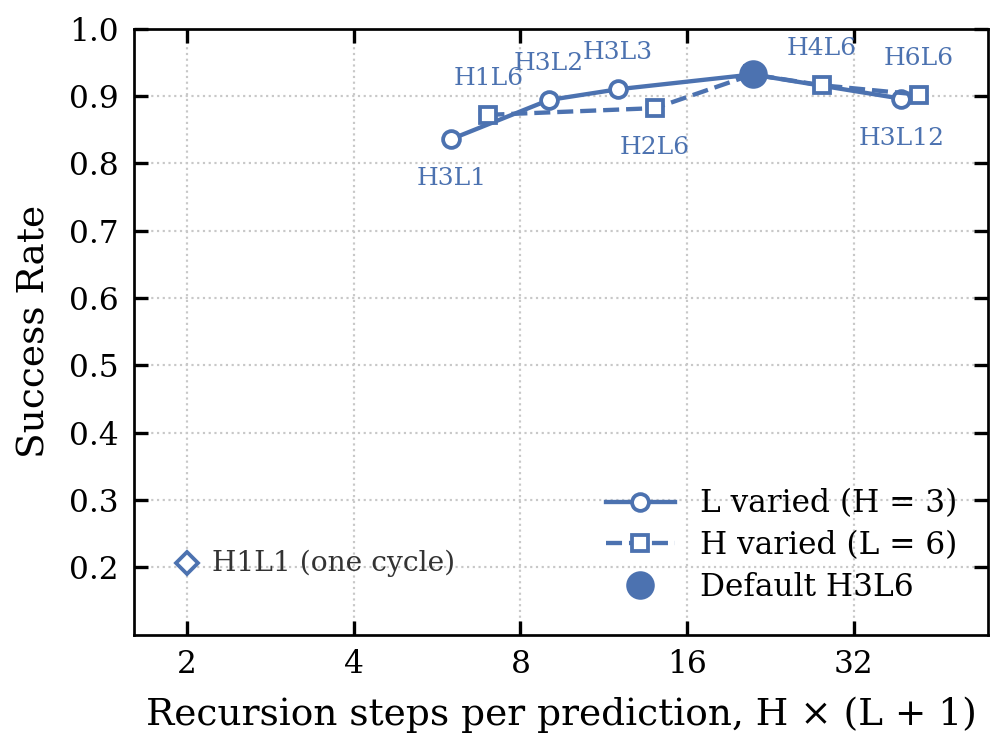}
        \caption{Recursion depth.}
        \label{fig:cycles}
    \end{subfigure}
    \caption{Ablations on LIBERO-LONG. (a) Success rate against model backbone parameter count. (b) Success rate against recursion steps
             per prediction}
    \label{fig:ablations}
\end{figure}

\paragraph{Interaction between carry and recursion.} Persistent state is not beneficial independently of recursive refinement. Table~\ref{tab:carry_recursion} shows a strong interaction between the two. With the default recursive configuration, enabling the carry substantially improves performance across all three benchmarks. At shallow recursion, however, the effect can reverse: on LIBERO-LONG, H1L1 drops from 0.642 without carry to 0.206 with carry, and MetaWorld shows the same qualitative pattern. LIBERO-Mem is the exception, where carry remains mildly beneficial even in the shallow model. These results suggest that persistent state must be sufficiently processed before it becomes useful. A shallow model may be unable to reconcile the incoming carry with the current observation, causing stale or poorly adapted latent state to interfere with action prediction. Additional recursive refinement may instead allow the policy to update, reinterpret, or overwrite the incoming state before decoding an action. We therefore do not view carry and recursion as independently beneficial components: the usefulness of persistent state depends strongly on having sufficient within-step computation to integrate it.

\paragraph{Temporal gradient propagation.} Table~\ref{tab:tbptt} tests whether the persistent carry should also transmit gradients between successive policy evaluations. The default TRP detaches the carry between timesteps, while the alternatives apply truncated backpropagation through time over increasingly long horizons. Propagating gradients through the carry consistently reduces performance on both LIBERO-LONG and LIBERO-Mem, with larger truncation horizons producing further degradation. Persistent forward state therefore provides the benefit observed by TRP without requiring long-range gradient propagation.

\paragraph{Model capacity.} Figure~\ref{fig:size} shows whether model capacity alone explains TRP's performance. Performance rises rapidly at small model sizes, with a 0.33M-parameter backbone already approaching the strongest much larger baselines on LIBERO-LONG. Performance peaks at the 4.2M-parameter configuration and then decreases as the backbone grows further. Thus, TRP's gains do not follow a simple scaling trend; recursively using a compact computation module appears more effective than increasing parameter count alone.

\paragraph{Recursion depth.} Finally, Figure~\ref{fig:cycles} varies the amount of recursive computation while holding the backbone fixed. Increasing either the number of high-level or low-level refinement cycles initially improves performance, with the default configuration achieving the strongest result. Additional recursion beyond this point provides no further benefit and eventually reduces performance. Together with the model-size ablation, this indicates that TRP benefits from additional computation through parameter reuse, but that the advantage saturates at a moderate recursion depth.

\section{Conclusion}
\label{sec:conclusion}

We introduced Tiny Recursive Policy, a stateful behaviour-cloning policy that combines recursive latent refinement with persistent temporal context for robotic manipulation. Rather than increasing model size or relying on iterative generative inference, TRP repeatedly reuses a compact computation module while propagating its latent state across successive policy evaluations. Across LIBERO, MetaWorld ML45, and LIBERO-Mem, TRP performs competitively in multi-task and few-shot settings and shows its strongest advantages on long-horizon and context-dependent manipulation tasks. Beyond aggregate performance, our execution-time interventions show that the role of the recurrent state depends strongly on the task. Wiping the carry has almost no effect on MetaWorld, but causes substantial degradation on LIBERO-LONG and LIBERO-Mem, indicating increasing reliance on persistent context as temporal dependencies become stronger. Training with carry perturbations substantially reduces this sensitivity and improves robustness to temporary visual blackout. Additional ablations show that persistent state and recursive refinement interact strongly, that propagating gradients through the carry is unnecessary, and that increasing recursion is more effective than simply increasing model capacity up to a moderate computation depth. Together, these results suggest that persistent state, when paired with sufficient recursive refinement, provides a parameter-efficient mechanism for building compact visuomotor policies for temporally demanding control.

\vspace{-2mm}
\paragraph{Limitations and Future Work}
While our results demonstrate strong performance, several limitations remain. First, our evaluation is currently restricted to simulation, and real-robot experiments are needed to establish whether the same benefits persist in real-world settings. Second, although our intervention experiments identify when the policy relies on its persistent carry, they do not reveal what information is encoded in the latent state, how individual task variables are represented, or how long particular information is retained. Probing or decoding the carry could more directly characterise the temporal representations learned by TRP. Finally, TRP currently uses a fixed amount of recursive computation for every prediction. Our recursion-depth experiments show that additional computation is beneficial only up to a point, motivating adaptive halting or input-dependent computation mechanisms that allocate refinement according to the difficulty of the current decision.

\bibliography{references}
\bibliographystyle{iclr2027_conference}

\end{document}